\documentclass{ecai}

\usepackage{latexsym}
\usepackage{amssymb}
\usepackage{amsmath}
\usepackage{amsthm}
\usepackage{booktabs}
\usepackage{enumitem}
\usepackage{graphicx}
\usepackage{color}
\usepackage{bm}
\usepackage{tabularx}
\usepackage{tikz}
\usepackage{adjustbox} 
\usepackage{xcolor}
\usepackage[ruled,vlined]{algorithm2e}
\usepackage{algorithmic}

\newcommand{\BibTeX}{B\kern-.05em{\sc i\kern-.025em b}\kern-.08em\TeX}

\begin{document}


\begin{frontmatter}




\title{ChatZero: Zero-shot Cross-Lingual Dialogue Generation via Pseudo-Target Language}



\author[1,2]{\fnms{Yongkang}~\snm{Liu}}
\author[1]{\fnms{Feng}~\snm{Shi}}
\author[1]{\fnms{Daling}~\snm{Wang}\thanks{Corresponding Author}} 
\author[1]{\fnms{Yifei}~\snm{Zhang}} 
\author[2]{\fnms{Hinrich}~\snm{Schütze}}

\address[1]{Northeastern University, China}
\address[2]{Center for Information and Language Processing and Munich Center for Machine Learning, LMU Munich}


\begin{abstract}
Although large language models(LLMs) show amazing capabilities, among various exciting applications discovered for LLMs fall short in other low-resource languages.
Besides, most existing methods depend on large-scale dialogue corpora and thus building systems for dialogue generation in a zero-shot scenario remains a considerable challenge. To address this challenge,
we propose a novel end-to-end zero-shot dialogue generation model ChatZero based on cross-lingual code-switching method. First, we construct code-switching language and pseudo-target language with placeholders. Then for cross-lingual semantic transfer, we employ unsupervised contrastive learning to minimize the semantics gap of the source language, code-switching language, and pseudo-target language that are mutually positive examples in the high dimensional semantic space.
Experiments on the multilingual DailyDialog and DSTC7-AVSD datasets demonstrate that ChatZero can achieve more than 90\% of the original performance under the zero-shot case compared to supervised learning, and achieve state-of-the-art performance compared with other baselines.
\end{abstract}

\end{frontmatter}


\section{Introduction}

Open domain dialogue generation techniques have achieved significant progress thanks to the availability of large-scale dialogue datasets. Particularly, the pre-trained large models of dialogue generation can generate informative and fluent responses~\cite{song2021bob,roller2021recipes}, which have huge potential in various applications such as emotional companionship, mental health support, and social chatbots.

The availability of large-scale datasets is a double-edged sword, which also brings a worrying phenomenon that most existing dialogue systems excessively rely on large-scale dialogue corpus~\cite{lewis2020bart,song2021bob,chen2022dialogved}. This phenomenon greatly limits the popularity of dialogue systems due to large-scale corpora unavailable in most cases~\cite{joshi2020state}. For example, there are more than 7,000 languages worldwide, but only about 1\% have an available corpus~\cite{wang2022expanding}. When it comes to dialogue tasks, there are even fewer languages with an available corpus.
Zero-shot dialogue techniques usually utilize non-target language corpus for knowledge transfer, 
which can significantly alleviate the dependence on the target language corpus~\cite{liu2022mulzdg}.

\begin{figure}[h]
\centering
\includegraphics[width=0.82\linewidth]{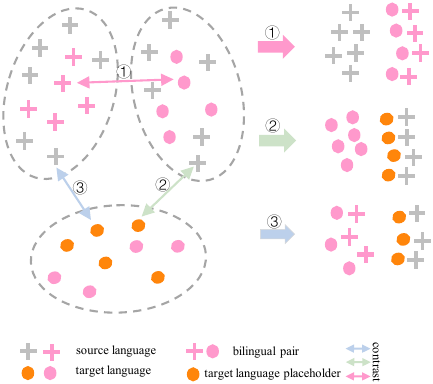}
\caption{Schematic diagram of ChatZero.}
\label{multi_semantic}
\end{figure}

The multilingual code-switching method has been proven to be effective in low- and zero-shot generation in NMT (Neural Machine Translation)~\cite{irvine2013combining,chen2022towards,kim2021model,lee2022pre}. Unfortunately, zero-shot generation methods in NMT are difficult to apply to dialogue tasks. The main reason is that the source and target languages of NMT have the same semantics, while there is no similar semantic phenomenon between dialogue history and response, as their semantics are different.

The most existing low-resource dialogue studies pay little attention to the problem of missing corpora. Low-resource knowledge-grounded dialogue generation methods focus on the problem of knowledge deficiency with sufficient corpus~\cite{li2020zero,zhao2020low,liu2021three}, which fail to work in zero-shot scenarios. ~\citet{liu2022mulzdg} proposes a utterance level code-switching method for zero-shot dialogue generation, which depends on massive translated target language utterances. Their proposed model exposes a large amount of target language corpus in exchange for performance improvement, which, strictly speaking, could not be regarded as a zero-shot generation. Although employing large-scale pre-trained language models, such as GPT-3~\cite{brown2020language}, BART~\cite{lewis2020bart}, and BlenderBot~\cite{roller2021recipes}, can reduce the dependence on target corpus, these models are usually limited to the language of pre-trained corpus and fail to work well on other languages.

It is known that different language representations of the same utterance are similar in high-dimensional semantic space~\cite{karlgren2019high}. Accordingly, we construct a pseudo-target language corresponding to the source language by dictionaries. The pseudo-target language refers to a language that contains target language words and placeholder \textbf{[MASK]}. For example, "\textit{Hier [MASK] ein [MASK]}"
is pseudo-German language. The main reason is that [MASK] can be considered as an unrevealed token with actual semantics in masked language model, such as mBERT, which is determined by the pre-training tasks.
Besides, we build code-switching languages consisting of source and target languages, through bilingual dictionaries to reduce the difficulty of cross-lingual learning. An example of code-switching between English and German is "\textit{Here is ein Beispiel}".
In summary, the semantics of the source language, pseudo-target language, and code-switching language of the same utterance are similar in semantic space.

We propose a novel end-to-end zero-shot dialogue generation model ChatZero based on a pseudo-target language. ChatZero employs unsupervised contrastive learning to minimize the representation gap of same utterances in different languages and maximize that of irrelevant utterances. As shown in Figure~\ref{multi_semantic}, we pull the semantic representations of the source language, the code-switching language and the pseudo-target language closer for same utterances. At the same time, we push away the utterances that are irrelevant in the same batch. 
Cross-lingual knowledge transfer is realized by semantic approximation in high-dimensional semantic space, which includes three aspects: (i) aligning the semantics between source and target languages (\textcolor{pink}{$\leftrightarrow$} and \textcolor{cyan!40!white}{$\leftrightarrow$}); (ii) aligning the semantics of the placeholders with the source language (\textcolor{green!40!white}{$\leftrightarrow$} and \textcolor{cyan!40!white}{$\leftrightarrow$}); (iii) aligning the semantics of the placeholders to the target language (\textcolor{cyan!40!white}{$\leftrightarrow$}). The process (iii) depends on processes (i) and (ii).
Because the input in the code-switching form contains limited vocabulary in the target language, (i) can only play a limited role in semantic transfer. The implicit semantic alignment between source and target languages is achieved through (ii) and (iii). Models can adapt to the input of the target language and improve the semantic transfer ability by (ii) and (iii).
These three aspects promote each other to transfer knowledge from the source to the target language. ChatZero allows placeholders to be included in the generated responses. Finally, we employ mBERT~\cite{kenton2019bert} to convert the placeholders into actual words.
To summarize, we make the following contributions:
\begin{itemize}
    \item We propose the idea of constructing a pseudo-target language by introducing placeholders. As far as we know, we are the first to study zero-shot dialogue generation task without using massive target language utterances.
    \item We propose a novel end-to-end zero-shot dialogue generation model ChatZero, which achieves cross-lingual knowledge transfer by minimizing representations in different languages through unsupervised contrastive learning.
    \item Extensive experiments on two multilingual benchmark datasets demonstrate that ChatZero can achieve more than 90\% of the original performance under zero-shot conditions compared to supervised learning and achieve state-of-the-art performance compared with other baselines. Code associated with our work are available on gitHub repository\footnote{https://github.com/misonsky/ChatZero}.
\end{itemize} 


\section{RELATED WORK}
\subsection{Dialogue Generation}
Dialogue generation systems aim to produce informative and fluent responses and have attracted considerable attention in academia. Early studies~\cite{sordoni2015hierarchical,serban2016building} employing seq2seq~\cite{sutskever2014sequence} structure tend to generate dull and generic responses. Since the emergence of the transformer~\cite{vaswani2017attention}, it has gradually become the go-to method. The popularity of Transformers brings a new problem of heavy reliance on large-scale corpus, such as DialogGPT~\cite{zhang2020dialogpt}, BlenderBot~\cite{roller2021recipes}, and LaMDA~\cite{thoppilan2022lamda}. Although they achieve promising performances, depending on large-scale corpora severely limits the usability of dialogue systems. Most languages that are recorded have no corpora available~\cite{joshi2020state,costa2022no,wang2022expanding}. This means that these methods fail to work in these languages. We propose a novel end-to-end zero-shot dialogue generation model to alleviate this problem.
\subsection{zero-shot Learning}

Dialogue generation has enjoyed a great boost utilizing neural network models. However, this is not the case for most languages, especially zero-shot ones with insufficient training corpus~\cite{chen2018zero,liu2024unified}.
Zero-shot learning is a method of learning without any target language training samples.
One of the solutions to zero-shot learning is cross-lingual transfer learning method, which improves the performance in the zero-shot target language by leveraging data from other (source) languages, typically with the help of cross-lingual resources. Cross-lingual transfer methods have been widely adopted in natural language processing tasks such as machine translation~\cite{johnson2017google,cheng2019joint,liu2020multilingual}. In this paper, we propose a cross-lingual, zero-shot generative model for dialogue generation, which does not depend on the target corpus.
\section{Problem Statement}
\subsection{Problem Formalization}
Given the source-language dialogue corpus and source-to-other language bilingual dictionaries, our goal is to build dialogue generation systems for other languages. In this paper, English is the source language. We concatenate the dialogue history of source language into a continuous sequence, denoted as $\bm{\tilde{H}}$, and the response denoted as $\bm{\tilde{R}}$. We employ $\bm{D}_t$ to denote a bilingual dictionary from English to target language \textit{t}.
\subsection{Code-switching Languages}
Constructing the code-switching language containing source and target language is a common method for cross-lingual semantic transfer. During the training process, input is provided in the form of code-switching, whereas during inference, the input is in the form of the target language. The gap makes models unable to adapt to zero-shot scenarios. On the other hand, the target language tokens included in the training is limited, resulting in poor semantic transfer ability of models. We propose constructing a pseudo-target code-switching language to alleviate this limitation.

We employ bilingual dictionaries\footnote{https://github.com/facebookresearch/MUSE\#ground-truth-bilingual-dictionaries} to build code-switching languages. The dictionary is English-centric, including En-Zh (English-Chinese), En-De (English-German), En-Ru (English-Russian), En-Es (English-Spanish), En-Fr (English-French), En-It (English-Italian). We also collect other bilingual dictionaries and expand the size of the dictionary to improve its coverage of the corpus.
The statistical information of the dictionaries is shown in Table~\ref{tab:dict}.
An English word may have multiple counterparts in other languages with the same meaning. 
At the same time, we count the coverage of different dictionaries in the English corpus. The calculation method is shown as follows:
\begin{equation}
f = \frac{Count(Dict \cap Corpus)}{Count(Corpus)}
\end{equation}
Count means no repeat count function. We remove stop words and punctuation.
The results are shown in Table~\ref{tab:dict}. It can be seen that the dictionaries have a high coverage rate for DSTC7-AVSD. The coverage rate has a direct impact on the performance of ChatZero.

Next, we introduce the process of constructing code-switching languages. We mainly build two forms of code-switching languages: (i) code-switching languages consisting of source and target languages; (ii) fake target languages containing placeholders and target languages. We employ the "[MASK]" symbol for placeholders. We can employ mBERT to manifest "[MASK]" a concrete token. We adopt Algorithm~\ref{alg:cs} to represent the construction process of code-switching languages.

\textbf{Symbolic descriptions}: The input of Algorithm~\ref{alg:cs} is $\bm{\tilde{H}}$, $\bm{\tilde{R}}$, and $\bm{D}$. $\bm{\tilde{H}}$, $\bm{\tilde{R}}$, where $\bm{\tilde{H}}$ and $\bm{\tilde{R}}$ are the source language.

We employ $\bm{\bar{H}=\{h_1,h_2,\cdots\}}$ and $\bm{\bar{R}=\{r_1,r_2,\cdots\}}$ to denote the dialogue histories and responses sets output by Algorithm~\ref{alg:cs}, where $\bm{h_i}=\{w_1,w_2,\cdots,w_s\}$ and $\bm{r_i}=\{w_1,w_2,\cdots,w_t\}$, $i$ represents the $i$-th example generated, $s$ and $t$ denote the sequence length of dialogue history and response, respectively. $\bm{\tau}$ represents a threshold value. The output $\bm{\bar{H}}$ and $\bm{\bar{R}}$ contain pseudo-target and code-switching lang. The $\bm{k}$ represents the number of iterations.
\begin{figure*}[t]
\centering 
\includegraphics[width=0.85\textwidth]{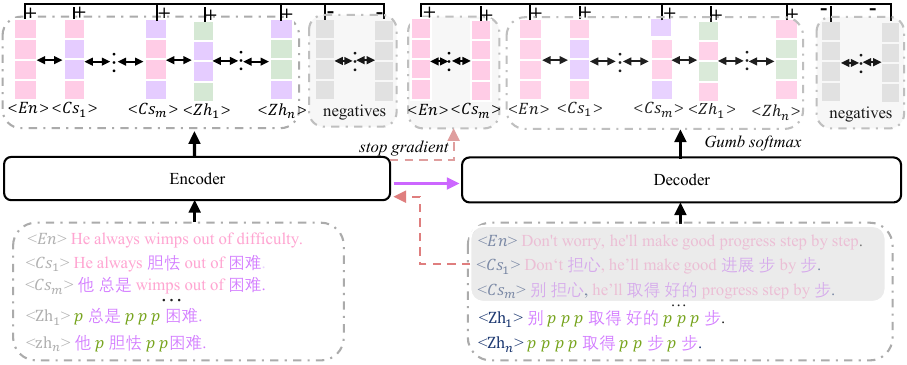}
\caption{Overview of ChatZero. $m$ represents the number of code-switching samples. $n$ represents the number of fake target language samples. $p$ means placeholder.}
\label{framework}
\end{figure*}

To distinguish different languages, we add an additional language identification token preceding both dialogue history and response. Specifically, English, Chinese, German, Spanish, French, Italian, and Russian are respectively set as: <En>, <Zh>, <De>, <Es>, <Fr>, <It> and <Ru>. In particular, we employ [Cs] as the language identification token for the code-switching language. The dialogue history and response of an English example can be expressed as $\bm{h}_{en}=\{$<En>$,w_1,w_2,\cdots,w_s\}$ and $\bm{r}_{en}=\{$<En>$,w_1,w_2,\cdots,w_t\}$. Adding language identification tokens for other languages is similar to English.
\begin{table}[ht]
\caption{Information of Dictionaries. \textbf{\#key} represents the number of key values in the dictionary, \textbf{\#val} represents the number of values, \textbf{\#cov-da} and \textbf{\#cov-ds} represent the dictionary's coverage of english training corpus on data sets DailyDialog and DSTC7-AVSD respectively.}
\begin{adjustbox}{max width=0.85\columnwidth, center}
\includegraphics[width=\textwidth]{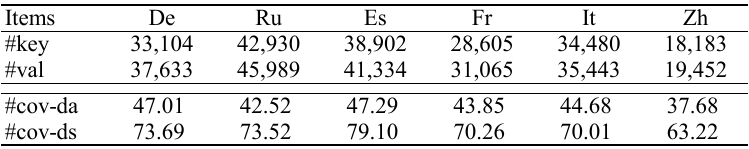}
\end{adjustbox}
\label{tab:dict}
\end{table}

\begin{algorithm}[t]
  \footnotesize
  \caption{Code-Switching Languages.}
  \label{alg:cs}
  \SetKwFunction{GetD}{GetDictValues}
  \SetKwFunction{TokenSelect}{RandGetTokens}
  \SetKwFunction{AddOp}{AddOperation}
  \SetKwFunction{UpdateSetOperation}{UpdateSetOperation}
  \SetKwFunction{ClearOperation}{ClearSetOperation}
  \SetKwFunction{RandomNum}{RandomNumber}
  \KwIn{history$ \bm{\tilde{H}}$; response $\bm{\tilde{R}}$; dictionary $D$; parameter $k$, $\tau$; placeholder symbol $S$.}

  \KwOut{code-switching history $\bm{\bar{H}}$ and response $\bm{\bar{R}}$}
  \BlankLine
  Initialize $\bm{\bar{H}}$, $\bm{\bar{R}}$ and local variable $\bm{h}$ as empty set $\emptyset$;

  \While{k>0}{

  \ForEach{token in $\bm{\tilde{H}}$ or $\bm{\tilde{R}}$}
  {

    \textbf{If} token in $D$ \textbf{do}

        \qquad tokens $\leftarrow$ \GetD{$D, token$}

        \qquad selection $\leftarrow$ \TokenSelect{tokens}

        \qquad \AddOp{$\bm{h}$,selection}

     \textbf{else}

        \qquad \AddOp{$\bm{h}$,$S$}

  }

  \UpdateSetOperation{$\bm{\bar{H}},\bm{\bar{R}},\bm{h}$}

  \ClearOperation{$h$}

  \ForEach{token in $\bm{\tilde{H}}$ or $\bm{\tilde{R}}$}
  {

    \textbf{If} token in $D$ \textbf{do}

        \qquad tokens $\leftarrow$ \GetD{$D, token$}

        \qquad \textbf{If} \RandomNum{} > $\tau$ \textbf{do}

                \qquad selection $\leftarrow$ \TokenSelect{tokens}

                \qquad \AddOp{$\bm{h}$,selection}

        \qquad \textbf{else}

            \qquad \AddOp{$\bm{h}$,tokens}

     \textbf{else}

        \qquad \AddOp{$\bm{h}$,$S$}

  }

  \UpdateSetOperation{$\bm{\bar{H}},\bm{\bar{R}},\bm{h}$}

  k = k-1
  }
\end{algorithm}

\section{METHODOLOGY}
The overall framework is illustrated in Figure~\ref{framework}. We use contrastive learning to minimize the semantic gaps between source language (i.e., English), code-switching language, and pseudo-target language at the encoding and decoding ends. ChatZero enhances the cross-lingual semantic transfer ability by direct and implicit semantic alignment. Besides, it also adapts to different forms of input during the inference stage.
\subsection{Multilingual Transformer}
A multilingual dialogue generation model learns a function $f$ to model the relation between dialogue history and response, which can be applied to different languages. Unfortunately, there are no pre-trained models available for multilingual dialogue generation. An alternative is to use the mBERT initialized Transformer as multilingual dialogue generation~\cite{rothe2020leveraging}. Specifically, the encoder and decoder are initialized by mBERT checkpoints.
\subsection{Cross-lingual Contrastive Learning}
Cross-lingual mechanisms enable the implicit learning of shared representations of different languages. ChatZero introduces contrastive loss to explicitly bring different languages together to map a shared semantic space. The core idea of contrastive learning is to minimize the representation gap of similar utterances and maximize that of irrelevant utterances. We leverage contrastive learning on the encoder side and decoder side, respectively.

We assume that the output at the encoder side is denoted as $\bm{\widetilde{h}}=\{\widetilde{h}_{cls},\widetilde{h}_{lan},\widetilde{h}_1,\widetilde{h}_2,\cdots,\widetilde{h}_s,\widetilde{h}_{sep}\}$, where $h_{lan}$ represents the representation of language identification token. The mean of all token representations is considered the representation of dialogue history, denoted as $c$, and the calculation method is as follows:
\begin{eqnarray}
 c = \frac{1}{s}\sum_i \widetilde{h}_i
\end{eqnarray}
According to Algorithm~\ref{alg:cs}, we will get 2$\times k$ +1 examples that are mutually positive instances. For 2$\times k$ +1 positive examples, we push their semantics close by maximizing the cosine similarity between them, which can be formally described as:
\begin{eqnarray}
 \ell_p^e = \frac{1}{2k+1}\sum_{i>j}^{2k+1} \frac{c_i^Tc_j}{||c_i||||c_j||}
\end{eqnarray}
where $\ell_p^e$ represents the similarity score of multiple positive examples. Besides, we maximize the distinction between positive and negative examples by maximizing the cosine similarity between each positive and negative pair.
Negative examples are other samples from the same batch. The loss is calculated as follows:
\begin{eqnarray}
 \ell_n^e = \frac{1}{2k+1}\sum_i^{2k+1} \sum_j\frac{c_i^TN_j}{||c_i||||N_j||}
\end{eqnarray}
where $N_j$ represents the $j$-th negative example in the same batch.

On the decoder side, we assume that the probabilities obtained by decoding is $\bm{P} \in \mathbb{R}^{t \times v}$, where $t$ is the decoded length and $v$ stands for the size of vocabulary $\bm{V}$.
We employ \textit{Gumb-Softmax} to sample the probability $\bm{P}$ to get the predicted probability distribution. The process can be formally described as follows:
\begin{eqnarray}
\tilde{\bm{P}} = \text{Gumb-Softmax}(\bm{P})
\end{eqnarray}
where $\tilde{\bm{P}} \in \mathbb{R}^{t \times v}$ and $\sum_j \tilde{\bm{P}}_{ij}$ = 1. $\tilde{\bm{P}}$ 
not only represents the probability distribution of each word appearing in the response, but also can be 
regarded as the weight corresponding to each word. Therefore, we can obtain the semantic representation of the predicted response through the dot product of $\tilde{\bm{P}}$ and $\bm{V}$. The process can be formally described as follows:
\begin{eqnarray}
\label{equa:six}
\bm{r} = \tilde{\bm{P}} \bm{V}
\end{eqnarray}
where $\bm{r} \in \mathbb{R}^{t \times d}$ and $d$ represents the hidden dimension of embedding layer. $\tilde{\bm{P}}$ assigns the correct candidate token a higher weight score, and the weight values of the remaining incorrect candidate tokens are close to zero.
Equation~\ref{equa:six} considers the information of incorrect tokens in a weighted approach, which mainly draws on the idea of label smoothing~\cite{muller2019does}.

Note that $\bm{r}$ contains placeholders [MASK]. The representation of response is denoted as $\tilde{\bm{r}}$ by taking the mean of all tokens representations. For predicted responses, we employ contrastive learning to minimize the gap between positive examples and maximize that between positive and negative examples. To minimize the number of placeholders in the predictions, we introduce the ground truth responses as rectification signals to the positive examples. We only adopt source language (i.e., English) and code-switching responses as rectification signals. The reason is that fake target responses contains placeholders, which will encourage model to generate responses with more [MASK]. We only obtain representations of ground truth responses with encoder and do not compute the gradients. On the decoder side, $\ell_p^d$ and $\ell_n^d$ are calculated as follows on decoder side:
\begin{eqnarray}
 \ell_p^d = \frac{1}{3k+2}\sum_{i>j}^{3k+2} \frac{\tilde{\bm{r}}_i^T\tilde{\bm{r}}_j}{||\tilde{\bm{r}}_i||||\tilde{\bm{r}}_j||}
\end{eqnarray}
\begin{eqnarray}
 \ell_n^d = \frac{1}{3k+2}\sum_i^{3k+2} \sum_j\frac{\tilde{\bm{r}}_i^TN_j}{||\tilde{\bm{r}}_i||||N_j||}
\end{eqnarray}
where $N_j$ represents the $j$-th negative example. We get the negative examples from other ground truth responses in the same batch. Besides, the loss for generative responses is cross-entropy, defined as:
\begin{eqnarray}
 \ell_g = \sum -log f_{\theta}(w_r|w_h)
\end{eqnarray}
The final total loss is defined as:
\begin{eqnarray}
L = \ell_g + \frac{1}{4t}(\ell_n^e + \ell_n^d - \ell_p^d - \ell_p^e)
\end{eqnarray}
Since $\ell_g$ is calculated on the token-level, therefore contrastive loss should be multiplied by the averaged response length $t$.
\section{Experiments}
\subsection{Datasets}
We employ multilingual versions of \textbf{DailyDialog} and \textbf{DSTC7-AVSD} datasets~\cite{liu2022mulzdg}, which include seven language versions i.e., English, Chinese, German, Russian, Spanish, French and Italian. The English version is the original corpus, and other language versions are obtained through bilingual dictionaries and translations. The supervised model for each language is trained on the corresponding language corpus.
\begin{table*}[ht]
\centering
\caption{Performance of the ChatZero comparison under zero-shot and supervised learning on DailyDialog. The \textbf{sup} and \textbf{zero} denote supervised learning performance and zero-shot performance, respectively. \textbf{Per} represents the percentage of zero-shot performance to supervised learning performance. \textbf{AVE} represents the average performance excluding PPL.}
\resizebox{0.78\textwidth}{!}{
\includegraphics[width=\textwidth]{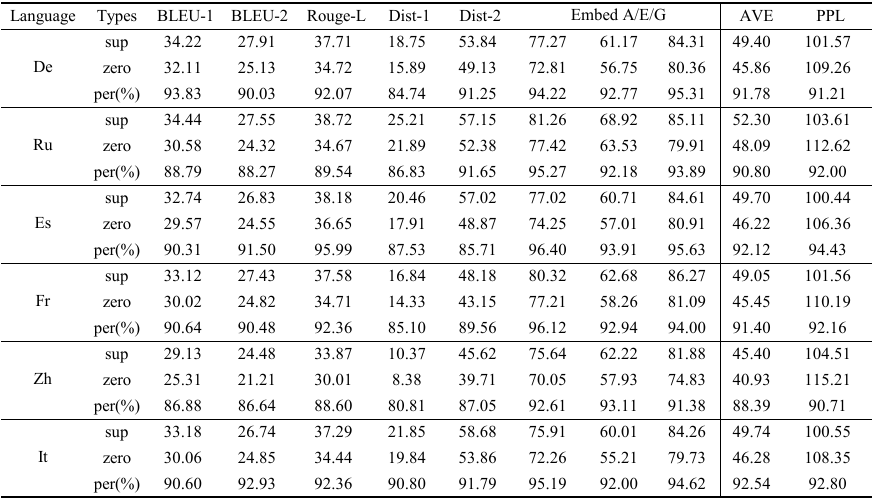}
}
\label{tab:dialy}
\end{table*}
\textbf{DailyDialog} is a multi-turn dialogue dataset of daily life, which consists of 11,118 context-response pairs for training, 1,000 pairs for validation, and 1,000 pairs for testing.
\textbf{DSTC7-AVSD} is a social media multi-turn dialogue dataset that consists of 76,590 context response pairs for training, 17,870 pairs for validation, and 1,710 pairs for testing.
\subsection{Baselines}
We compare the proposed model performance with the following baselines:
\textbf{LVM}~\cite{liu2019zero} which refines the aligned cross-lingual word-level representations by very few parallel word pairs. In this paper, we refine the cross-lingual word-level representations by bilingual dictionaries.
\textbf{MLT}~\cite{liu2020attention} which leverages parallel word pairs to generate code-switching sentences for learning the interlingual semantics across languages
\textbf{OBPE}~\cite{patil2022overlap} which modifies the BPE algorithm to encourage more shared tokens between high-resource and low-resource languages tokens in the vocabulary. We obtain the OBPE word vocabulary based on bilingual dictionaries. We use mBERT as the base model and use mBERT tokenizer. This method does not work for source and target languages that have a big gap. Therefore, we also consider all source and target token pairs that appear in bilingual dictionaries share the same representation.
\subsection{Implementation Details}
We implement our ChatZero using PyTorch and train ChatZero on a server with an Intel(R) Xeon(R) Gold-5218R CPU 2.10GHz and 4×GeForce RTX 3090 GPU (24G). Adam~\cite{kingma2014adam} is utilized for optimization. The adam parameters beta1 and beta2 are set to 0.9 and 0.999, respectively. Note that we employ mBERT to perform placeholder (i.e., [MASK]) reprediction of ChatZero results. When predicting [MASK], we will concatenate the dialogue history and generated responses with [MASK] into a continuous sequence and input it into mBERT.
The maximum length of the dialogue history is set to 512, and the maximum length of the response set to 50. We set the batch size to 64 and the learning rate to 5e$^{-5}$. Beam search is used to generate responses. The beam size is set 6. We train word embeddings for embedding-based metrics for each language using Glove~\cite{pennington2014glove}. The unknown tokens are removed when computing embedding-based metrics, and the vectors of all unknown tokens are initialized to zero vector. We find that the parameter $k$ is set to 2 for the best performance of ChatZero under zero-shot condition. The main reason is that most bilingual dictionaries have more than two candidate replacement words in target language with a ratio of 50\%. Based on validation set experiments, the parameter $\tau$ is set to 0.4.
\begin{table*}[ht]
\centering
\caption{Performance comparison under zero-shot and supervised learning on DSTC7-AVSD.}
\resizebox{0.78\textwidth}{!}{
\includegraphics[width=\textwidth]{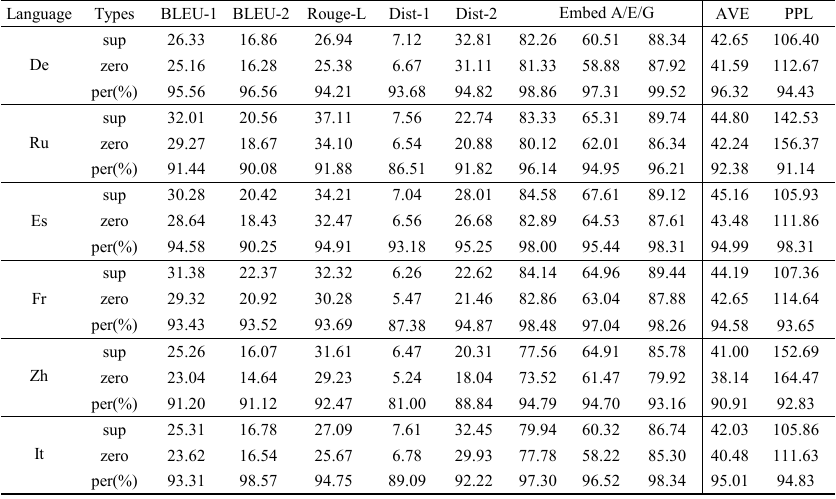}
}
\label{tab:dstc}
\end{table*}
\begin{table*}[ht]
\centering
\caption{Performances of baselines comparison on DailyDialog (up) and DSTC7-AVSD (down). \textbf{Bold} indicates the best result, and \underline{underline} indicates the second best result.}
\resizebox{0.78\textwidth}{0.43\textheight}{
\includegraphics[width=\textwidth]{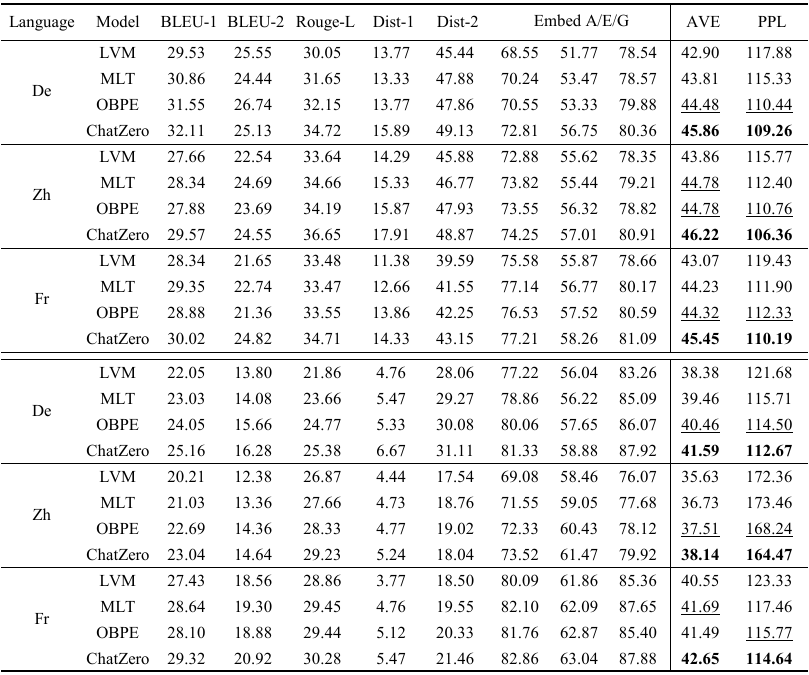}
}
\label{tab:base}
\end{table*}

\subsection{Evaluation metrics}
We employ both automatic metrics and human evaluations. \textbf{Automatic Metrics}: Following previous studies~\cite{song2021bob,liu2022mulzdg}, we employ perplexity (PPL) and distinct-1/2 (i.e., Dist.1/2). A lower PPL means a more reliable result. Distinct-1/2 is a key metric to evaluate the diversity of responses, which can be calculated through the ratio of distinct uni-grams / bi-grams. Higher distinct means better diversity of responses generated by the model. The higher the diversity of responses generated by the model, the higher the value of the Distinct-1/2 metrics. Following previous studies~\cite{song2021bob,liu2022mulzdg,liu2022pvgru}, we also employ BLEU and ROUGE-L for evaluating response generation. BLEU~\cite{chen2014systematic} and ROUGE-L~\cite{chen2014systematic} metrics evaluate the response based on co-occurrence properties of tokens. Embedding-based metrics (Average, Exterma, and Greedy)~\cite{liu2016not,xu2018better,sedoc2019chateval} can reflect the quality of the generated responses at the semantic level.

\textbf{Human Evaluation}: Human evaluation mainly includes the following three aspects: (i) \textbf{Fluency} measures whether the generated responses are smooth or grammatically correct. (ii) \textbf{Diversity} evaluates whether the generated responses are informative, rather than generic and repeated  information. (iii) \textbf{Relevance} evaluates whether the generated responses are relevant to the dialogue context. We select Chinese, French and German responses for human evaluation. We ask three crowdsourced graduate students to evaluate the quality of generated responses for 100 randomly sampled input contexts. We request annotators to score the response quality on a scale of [0,1,2] (0-bad, 1-neutral, 2-good) from three aspects fluency, diversity, and relevance. All annotators are unaware of the model corresponding to the generated results.

\subsection{Results}
Table~\ref{tab:dialy} and Table~\ref{tab:dstc} report the results of automatic metrics of ChatZero on DailyDialog and DSTC7-AVSD datasets under supervised and zero-shot conditions, respectively. First, ChatZero's performance under zero-shot conditions is inferior to that of supervised learning. However, the performance of zero-shot is exciting. Specifically, apart from Chinese datasets, ChatZero's zero-shot overall performance surpasses 90\% of that achieved through supervised learning on DailyDialog.  
On DSTC7-AVSD dataset, ChatZero's overall zero-shot performance exceeds 90\% of supervised learning across all languages. The results demonstrate the effectiveness of ChatZero. 

We find an interesting phenomenon that ChatZero's cross-lingual capabilities are relatively weaker in Chinese and Russian compared to other languages. Specifically, ChatZero's zero-shot performance in Chinese is only 88.39\% of supervised learning, and 90.8\% in Russian on DailyDialog. On DSTC7-AVSD, ChatZero's zero-shot performance in Chinese is only 90.91\% of supervised learning, and 92.38\%in Russian. We believe this is related to the dictionary's coverage of the corpus and the similarity between languages. The corresponding discussion is in Section~\ref{cross-analy}. We can also observe similar phenomena on different metrics.
On DailyDialog, the zero-shot BLEU-1/2 and Rouge-L results of ChatZero can achieve the performance of more than 90\% of supervised learning in German, Spanish, French, and Italian. The performance of zero-shot learning on PPL and Embedding-based metrics have achieved more than 90\% of supervised learning on all tested languages. On DSTC7-AVSD, we can observe that, except for the Distinct-1/2 metrics, the performance of other metrics under zero-shot reaches more than 91\% of supervised learning. The results of some metrics under zero-shot are even close to that of supervised learning.

Table~\ref{tab:base} reports the performance of ChatZero and other baselines on two datasets. We can observe that ChatZero enjoys the advantage of performance compared to other baselines. LVM achieves cross-lingual semantic transfer by training a shared embedding on word pairs. The disadvantage of LVM is that the size and coverage of word pairs strictly limit the model's semantic transfer capabilities, which may cause models to appear OOV (out-of-vocabulary) on the target language. MLT achieves cross-semantic transfer by constructing code-switching sentences through corpus pairs. The inability to fully cover the target language often results in the generation of code-switching results. OBPE also constructs co-embedding between high-resource and low-resource languages, and faces the same problem as LVM. ChatZero avoids the challenge of generating code-switching forms by creating a pseudo-target language structure. It fills in placeholders in a manner that aligns with the language model's pre-training, compensating for the limited coverage of dictionaries.
\subsection{Cross-lingual Analysis} 
\label{cross-analy}
We can observe that ChatZero demonstrates different cross-lingual abilities in different languages. The Chinese zero-shot performances of ChatZero on DailyDialog and DSTC7-AVSD have a gap compared to supervised learning than other languages. We believe this is mainly related to dictionary coverage and language similarity. From the dictionary coverage, the English-Chinese dictionary coverage of DailyDialog is 37.68\%, and DSTC7-AVSD is 63.22\%, which is the lowest compared to other dictionaries. The higher the dictionary coverage, the higher the bilingual pairs contained in the training corpus, which will significantly enhance the ability of ChatZero to transfer knowledge from the source language to the target language. 

In order to explore the impact of language similarity on cross-lingual transfer learning, we calculate the similarity between English and other languages following previous study~\cite{nie2023cross}, we calculate the similarity between English and other languages, where Zh-En is 9.4\%, Ru-En is 41.01\%, Es-En is 54.51\%, Fr-En is 54.67\%, It-En is 68.65\% and De-En is 91.84\%. We find that Chinese and Russian are less similar to English than other languages, especially Chinese, align with our experimental observations, that is, ChatZero's cross-lingual capabilities are relatively weaker in Chinese and Russian compared to other languages. The similarity between languages can affect the cross-lingual ability of the model. We believe that cross-language transfer learning is easier between similar languages.

\begin{table}[ht]
\caption{Human evaluation results on DailyDialog (left) and DSTC7-AVSD (right).}
\begin{adjustbox}{max width=0.98\linewidth, center}
\includegraphics[width=\textwidth]{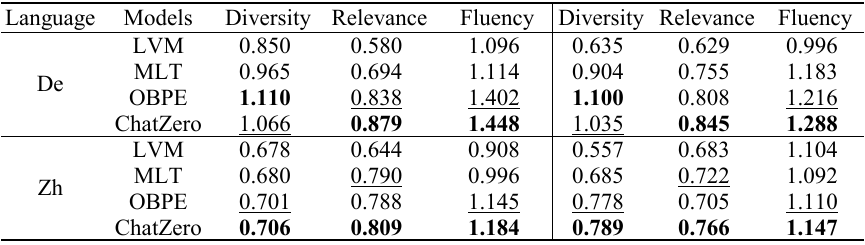}
\end{adjustbox}
\label{tab:human_evaluation}
\end{table}

\subsection{Human Evaluation \& Analysis}
We compare the performance of ChatZero with other baselines from three dimensions through human evaluation.
Table~\ref{tab:human_evaluation} reports the results. We can observe that ChatZero has obvious performance advantages in most evaluation dimensions. Specifically, on the German DailyDialog dataset,
ChatZero has 30.8\% advantage in diversity, 25.8\% in relevance and 32.2\% in fluency compared with LVM on DailyDialog. Compared with MLT, ChatZero has 11.6\% advantage in diversity, 13.8\% in relevance and 22.20\% in fluency. On the DSTC7-AVSD dataset, we can observe a similar phenomenon. As we discussed previously, LVM suffers from OOV problems on the target language and MLT generates code-switching responses, which is the main reason for the performance inferiority of these models. Although the performance of model OBPE is close to that of model ChatZero, it still has a slight disadvantage. The idea of OBPE maximizes word overlap is limited by the degree of similarity between source and target languages. This method is more effective when the source and target language are similar. We can observe that the performance of OBPE in German is closer to ChatZero than in Chinese, and even exceeds ChatZero in the diversity dimension.

To further evaluate the effectiveness of ChatZero, we count the proportion of placeholders in the responses generated by ChatZero. The ratio of placeholders [MASK] is obtained by dividing the number of placeholders by the number of all words in the responses.
The placeholders in the generated responses are at a low level in both datasets, with an average of 8.74\% placeholders on DailyDialog and 6.28\% on DSTC7-AVSD. We find that the placeholders of the generated responses on DailyDialog are significantly higher than those on DSTC7-AVSD, which shows that the higher the dictionary coverage, the lower the placeholders in the generated responses.

\begin{figure}[t]
\centering 
\includegraphics[width=0.99\linewidth]{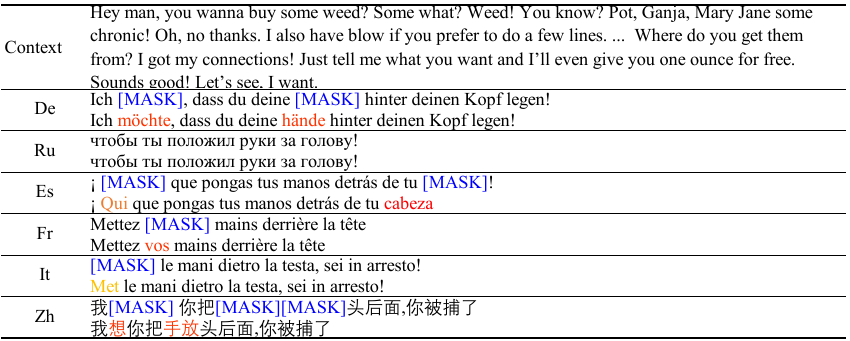}
\caption{The \textbf{context} stands for dialogue history. Here we only give the dialogue context in English. \textbf{De} represents the response in German under the zero resource condition, where the first row represents the response generated by ChatZero with placeholders, and the second row represents the responses of using mBERT to re-predict the placeholders. Other languages are similar. We highlight placeholders in \textcolor{blue}{blue}, correct predictions by mBERT in \textcolor{red}{red}, and incomplete words in \textcolor{orange}{orange}.}
\label{case1}
\end{figure}
\subsection{Case Study}
We further analyze the performance of the model through case Figure~\ref{case1}. It is an effective method to employ mBERT to predict the placeholders [MASK] when the generated responses contain fewer placeholders. We can observe that our approach still has some problems. The responses contains incomplete tokens. The main reason is that the number of tokens used to express the same semantics in different languages is inconsistent. We can not pre-set the number of placeholders when constructing pseudo-target language containing placeholders, which encourages ChatZero to assume that the number of tokens to express the same semantics is consistent in different languages. On the other hand, some words are splited into smaller tokens after word segmentation by the wordpiece tokenizer used in mBERT. These two reasons result in incomplete words being generated when making predictions for placeholders by mBERT.

\section{Conclusion}
We propose a novel zero-shot dialogue generation model ChatZero by introducing placeholders to build a pseudo-target language, which can avoid generating code-switching responses. ChatZero makes full use of the advantages of language models to make up for the shortcomings of incomplete dictionary coverage. Specifically,
ChatZero utilizes unsupervised contrastive learning to minimize the semantic gap of source, code-switching and pseudo-target languages. Results on two multilingual dialogue datasets show that ChatZero achieves more than 90\% of the supervised learning performance. Compared with baselines, ChatZero achieves state-of-the-art performance.
\section*{Acknowledgement}
We would like to thank reviewers for their constructive comments. The project is supported by the National Natural Science Foundation of China (62172086,62272092) and DFG (grant SCHU 2246/14-1). The project is also supported by China Scholarship Council.

\bibliography{mybibfile}

\begin{thebibliography}{42}
\providecommand{\natexlab}[1]{#1}
\providecommand{\url}[1]{\texttt{#1}}
\expandafter\ifx\csname urlstyle\endcsname\relax
  \providecommand{\doi}[1]{doi: #1}\else
  \providecommand{\doi}{doi: \begingroup \urlstyle{rm}\Url}\fi

\bibitem[Brown et~al.(2020)Brown, Mann, Ryder, Subbiah, Kaplan, Dhariwal,
  Neelakantan, Shyam, Sastry, Askell, et~al.]{brown2020language}
T.~Brown, B.~Mann, N.~Ryder, M.~Subbiah, J.~D. Kaplan, P.~Dhariwal,
  A.~Neelakantan, P.~Shyam, G.~Sastry, A.~Askell, et~al.
\newblock Language models are few-shot learners.
\newblock \emph{Advances in neural information processing systems},
  33:\penalty0 1877--1901, 2020.

\bibitem[Chen and Cherry(2014)]{chen2014systematic}
B.~Chen and C.~Cherry.
\newblock A systematic comparison of smoothing techniques for sentence-level
  bleu.
\newblock In \emph{Proceedings of the ninth workshop on statistical machine
  translation}, pages 362--367, 2014.

\bibitem[Chen et~al.(2022{\natexlab{a}})Chen, Ma, Chen, Zhang, Pan, Wang, and
  Wei]{chen2022towards}
G.~Chen, S.~Ma, Y.~Chen, D.~Zhang, J.~Pan, W.~Wang, and F.~Wei.
\newblock Towards making the most of cross-lingual transfer for zero-shot
  neural machine translation.
\newblock In \emph{Proceedings of the 60th Annual Meeting of the Association
  for Computational Linguistics (Volume 1: Long Papers)}, pages 142--157,
  2022{\natexlab{a}}.

\bibitem[Chen et~al.(2022{\natexlab{b}})Chen, Gong, Wang, Yao, Qi, Wei, Hu,
  Zhou, Mao, Chen, et~al.]{chen2022dialogved}
W.~Chen, Y.~Gong, S.~Wang, B.~Yao, W.~Qi, Z.~Wei, X.~Hu, B.~Zhou, Y.~Mao,
  W.~Chen, et~al.
\newblock Dialogved: A pre-trained latent variable encoder-decoder model for
  dialog response generation.
\newblock In \emph{Proceedings of the 60th Annual Meeting of the Association
  for Computational Linguistics (Volume 1: Long Papers)}, pages 4852--4864,
  2022{\natexlab{b}}.

\bibitem[Chen et~al.(2018)Chen, Awadallah, Hassan, Wang, and
  Cardie]{chen2018zero}
X.~Chen, A.~H. Awadallah, H.~Hassan, W.~Wang, and C.~Cardie.
\newblock Zero-resource multilingual model transfer: Learning what to share.
\newblock 2018.

\bibitem[Cheng(2019)]{cheng2019joint}
Y.~Cheng.
\newblock Joint training for pivot-based neural machine translation.
\newblock In \emph{Joint Training for Neural Machine Translation}, pages
  41--54. Springer, 2019.

\bibitem[Costa-juss{\`a} et~al.(2022)Costa-juss{\`a}, Cross, {\c{C}}elebi,
  Elbayad, Heafield, Heffernan, Kalbassi, Lam, Licht, Maillard,
  et~al.]{costa2022no}
M.~R. Costa-juss{\`a}, J.~Cross, O.~{\c{C}}elebi, M.~Elbayad, K.~Heafield,
  K.~Heffernan, E.~Kalbassi, J.~Lam, D.~Licht, J.~Maillard, et~al.
\newblock No language left behind: Scaling human-centered machine translation.
\newblock \emph{arXiv preprint arXiv:2207.04672}, 2022.

\bibitem[Devlin et~al.(2019)Devlin, Chang, Lee, and Toutanova]{kenton2019bert}
J.~Devlin, M.-W. Chang, K.~Lee, and K.~Toutanova.
\newblock Bert: Pre-training of deep bidirectional transformers for language
  understanding.
\newblock \emph{ArXiv}, abs/1810.04805, 2019.

\bibitem[Irvine and Callison-Burch(2013)]{irvine2013combining}
A.~Irvine and C.~Callison-Burch.
\newblock Combining bilingual and comparable corpora for low resource machine
  translation.
\newblock In \emph{Proceedings of the eighth workshop on statistical machine
  translation}, pages 262--270, 2013.

\bibitem[Johnson et~al.(2017)Johnson, Schuster, Le, Krikun, Wu, Chen, Thorat,
  Vi{\'e}gas, Wattenberg, Corrado, et~al.]{johnson2017google}
M.~Johnson, M.~Schuster, Q.~V. Le, M.~Krikun, Y.~Wu, Z.~Chen, N.~Thorat,
  F.~Vi{\'e}gas, M.~Wattenberg, G.~Corrado, et~al.
\newblock Google’s multilingual neural machine translation system: Enabling
  zero-shot translation.
\newblock \emph{Transactions of the Association for Computational Linguistics},
  5:\penalty0 339--351, 2017.

\bibitem[Joshi et~al.(2020)Joshi, Santy, Budhiraja, Bali, and
  Choudhury]{joshi2020state}
P.~Joshi, S.~Santy, A.~Budhiraja, K.~Bali, and M.~Choudhury.
\newblock The state and fate of linguistic diversity and inclusion in the nlp
  world.
\newblock In \emph{Proceedings of the 58th Annual Meeting of the Association
  for Computational Linguistics}, pages 6282--6293, 2020.

\bibitem[Karlgren and Kanerva(2019)]{karlgren2019high}
J.~Karlgren and P.~Kanerva.
\newblock High-dimensional distributed semantic spaces for utterances.
\newblock \emph{Natural Language Engineering}, 25\penalty0 (4), 2019.

\bibitem[Kim et~al.(2021)Kim, Jang, Jung, and Shin]{kim2021model}
S.~Kim, J.~Y. Jang, M.~Jung, and S.~Shin.
\newblock A model of cross-lingual knowledge-grounded response generation for
  open-domain dialogue systems.
\newblock In \emph{Findings of the Association for Computational Linguistics:
  EMNLP 2021}, pages 352--365, 2021.

\bibitem[Kingma and Ba(2014)]{kingma2014adam}
D.~P. Kingma and J.~Ba.
\newblock Adam: A method for stochastic optimization.
\newblock \emph{arXiv preprint arXiv:1412.6980}, 2014.

\bibitem[Lee et~al.(2022)Lee, Thillainathan, Nayak, Ranathunga, Adelani, Su,
  and McCarthy]{lee2022pre}
E.-S. Lee, S.~Thillainathan, S.~Nayak, S.~Ranathunga, D.~Adelani, R.~Su, and
  A.~D. McCarthy.
\newblock Pre-trained multilingual sequence-to-sequence models: A hope for
  low-resource language translation?
\newblock In \emph{Findings of the Association for Computational Linguistics:
  ACL 2022}, pages 58--67, 2022.

\bibitem[Lewis et~al.(2020)Lewis, Liu, Goyal, Ghazvininejad, Mohamed, Levy,
  Stoyanov, and Zettlemoyer]{lewis2020bart}
M.~Lewis, Y.~Liu, N.~Goyal, M.~Ghazvininejad, A.~Mohamed, O.~Levy, V.~Stoyanov,
  and L.~Zettlemoyer.
\newblock Bart: Denoising sequence-to-sequence pre-training for natural
  language generation, translation, and comprehension.
\newblock In \emph{Proceedings of the 58th Annual Meeting of the Association
  for Computational Linguistics}, pages 7871--7880, 2020.

\bibitem[Li et~al.(2020)Li, Xu, Wu, Zhao, Zhao, and Tao]{li2020zero}
L.~Li, C.~Xu, W.~Wu, Y.~Zhao, X.~Zhao, and C.~Tao.
\newblock Zero-resource knowledge-grounded dialogue generation.
\newblock \emph{Advances in Neural Information Processing Systems},
  33:\penalty0 8475--8485, 2020.

\bibitem[Liu et~al.(2016)Liu, Lowe, Serban, Noseworthy, Charlin, and
  Pineau]{liu2016not}
C.-W. Liu, R.~Lowe, I.~V. Serban, M.~Noseworthy, L.~Charlin, and J.~Pineau.
\newblock How not to evaluate your dialogue system: An empirical study of
  unsupervised evaluation metrics for dialogue response generation.
\newblock In \emph{Proceedings of the 2016 Conference on Empirical Methods in
  Natural Language Processing}, pages 2122--2132, 2016.

\bibitem[Liu et~al.(2021)Liu, Zhao, Li, Ren, Zhang, and Yin]{liu2021three}
S.~Liu, X.~Zhao, B.~Li, F.~Ren, L.~Zhang, and S.~Yin.
\newblock A three-stage learning framework for low-resource knowledge-grounded
  dialogue generation.
\newblock In \emph{Proceedings of the 2021 Conference on Empirical Methods in
  Natural Language Processing}, pages 2262--2272, 2021.

\bibitem[Liu et~al.(2020{\natexlab{a}})Liu, Gu, Goyal, Li, Edunov,
  Ghazvininejad, Lewis, and Zettlemoyer]{liu2020multilingual}
Y.~Liu, J.~Gu, N.~Goyal, X.~Li, S.~Edunov, M.~Ghazvininejad, M.~Lewis, and
  L.~Zettlemoyer.
\newblock Multilingual denoising pre-training for neural machine translation.
\newblock \emph{Transactions of the Association for Computational Linguistics},
  8:\penalty0 726--742, 2020{\natexlab{a}}.

\bibitem[Liu et~al.(2022{\natexlab{a}})Liu, Feng, Wang, Sch{\"u}tze, and
  Zhang]{liu2022pvgru}
Y.~Liu, S.~Feng, D.~Wang, H.~Sch{\"u}tze, and Y.~Zhang.
\newblock Pvgru: Generating diverse and relevant dialogue responses via
  pseudo-variational mechanism.
\newblock \emph{arXiv preprint arXiv:2212.09086}, 2022{\natexlab{a}}.

\bibitem[Liu et~al.(2022{\natexlab{b}})Liu, Feng, Wang, and
  Zhang]{liu2022mulzdg}
Y.~Liu, S.~Feng, D.~Wang, and Y.~Zhang.
\newblock Mulzdg: Multilingual code-switching framework for zero-shot dialogue
  generation, 2022{\natexlab{b}}.

\bibitem[Liu et~al.(2024)Liu, Nie, Hua, Ding, Wang, Zhang, and
  Sch{\"u}tze]{liu2024unified}
Y.~Liu, E.~Nie, Z.~Hua, Z.~Ding, D.~Wang, Y.~Zhang, and H.~Sch{\"u}tze.
\newblock A unified data augmentation framework for low-resource multi-domain
  dialogue generation.
\newblock \emph{arXiv preprint arXiv:2406.09881}, 2024.

\bibitem[Liu et~al.(2019)Liu, Shin, Xu, Winata, Xu, Madotto, and
  Fung]{liu2019zero}
Z.~Liu, J.~Shin, Y.~Xu, G.~I. Winata, P.~Xu, A.~Madotto, and P.~Fung.
\newblock Zero-shot cross-lingual dialogue systems with transferable latent
  variables.
\newblock In \emph{Proceedings of the 2019 Conference on Empirical Methods in
  Natural Language Processing and the 9th International Joint Conference on
  Natural Language Processing (EMNLP-IJCNLP)}, pages 1297--1303, 2019.

\bibitem[Liu et~al.(2020{\natexlab{b}})Liu, Winata, Lin, Xu, and
  Fung]{liu2020attention}
Z.~Liu, G.~I. Winata, Z.~Lin, P.~Xu, and P.~Fung.
\newblock Attention-informed mixed-language training for zero-shot
  cross-lingual task-oriented dialogue systems.
\newblock In \emph{Proceedings of the AAAI Conference on Artificial
  Intelligence}, volume~34, pages 8433--8440, 2020{\natexlab{b}}.

\bibitem[M{\"u}ller et~al.(2019)M{\"u}ller, Kornblith, and
  Hinton]{muller2019does}
R.~M{\"u}ller, S.~Kornblith, and G.~E. Hinton.
\newblock When does label smoothing help?
\newblock \emph{Advances in neural information processing systems}, 32, 2019.

\bibitem[Nie et~al.(2023)Nie, Liang, Schmid, and Sch{\"u}tze]{nie2023cross}
E.~Nie, S.~Liang, H.~Schmid, and H.~Sch{\"u}tze.
\newblock Cross-lingual retrieval augmented prompt for low-resource languages.
\newblock In \emph{The 61st Annual Meeting Of The Association For Computational
  Linguistics}, 2023.

\bibitem[Patil et~al.(2022)Patil, Talukdar, and Sarawagi]{patil2022overlap}
V.~Patil, P.~Talukdar, and S.~Sarawagi.
\newblock Overlap-based vocabulary generation improves cross-lingual transfer
  among related languages.
\newblock In \emph{Proceedings of the 60th Annual Meeting of the Association
  for Computational Linguistics (Volume 1: Long Papers)}, pages 219--233, 2022.

\bibitem[Pennington et~al.(2014)Pennington, Socher, and
  Manning]{pennington2014glove}
J.~Pennington, R.~Socher, and C.~D. Manning.
\newblock Glove: Global vectors for word representation.
\newblock In \emph{Proceedings of the 2014 conference on empirical methods in
  natural language processing (EMNLP)}, pages 1532--1543, 2014.

\bibitem[Roller et~al.(2021)Roller, Dinan, Goyal, Ju, Williamson, Liu, Xu, Ott,
  Smith, Boureau, et~al.]{roller2021recipes}
S.~Roller, E.~Dinan, N.~Goyal, D.~Ju, M.~Williamson, Y.~Liu, J.~Xu, M.~Ott,
  E.~M. Smith, Y.-L. Boureau, et~al.
\newblock Recipes for building an open-domain chatbot.
\newblock In \emph{Proceedings of the 16th Conference of the European Chapter
  of the Association for Computational Linguistics: Main Volume}, pages
  300--325, 2021.

\bibitem[Rothe et~al.(2020)Rothe, Narayan, and Severyn]{rothe2020leveraging}
S.~Rothe, S.~Narayan, and A.~Severyn.
\newblock Leveraging pre-trained checkpoints for sequence generation tasks.
\newblock \emph{Transactions of the Association for Computational Linguistics},
  8:\penalty0 264--280, 2020.

\bibitem[Sedoc et~al.(2019)Sedoc, Ippolito, Kirubarajan, Thirani, Ungar, and
  Callison-Burch]{sedoc2019chateval}
J.~Sedoc, D.~Ippolito, A.~Kirubarajan, J.~Thirani, L.~Ungar, and
  C.~Callison-Burch.
\newblock Chateval: A tool for chatbot evaluation.
\newblock In \emph{Proceedings of the 2019 conference of the North American
  chapter of the association for computational linguistics (demonstrations)},
  pages 60--65, 2019.

\bibitem[Serban et~al.(2016)Serban, Sordoni, Bengio, Courville, and
  Pineau]{serban2016building}
I.~Serban, A.~Sordoni, Y.~Bengio, A.~Courville, and J.~Pineau.
\newblock Building end-to-end dialogue systems using generative hierarchical
  neural network models.
\newblock In \emph{Proceedings of the AAAI Conference on Artificial
  Intelligence}, volume~30, 2016.

\bibitem[Song et~al.(2021)Song, Wang, Zhang, Zhang, and Liu]{song2021bob}
H.~Song, Y.~Wang, K.~Zhang, W.~Zhang, and T.~Liu.
\newblock Bob: Bert over bert for training persona-based dialogue models from
  limited personalized data.
\newblock In \emph{Proceedings of the 59th Annual Meeting of the Association
  for Computational Linguistics and the 11th International Joint Conference on
  Natural Language Processing (Volume 1: Long Papers)}, pages 167--177, 2021.

\bibitem[Sordoni et~al.(2015)Sordoni, Bengio, Vahabi, Lioma, Grue~Simonsen, and
  Nie]{sordoni2015hierarchical}
A.~Sordoni, Y.~Bengio, H.~Vahabi, C.~Lioma, J.~Grue~Simonsen, and J.-Y. Nie.
\newblock A hierarchical recurrent encoder-decoder for generative context-aware
  query suggestion.
\newblock In \emph{proceedings of the 24th ACM international on conference on
  information and knowledge management}, pages 553--562, 2015.

\bibitem[Sutskever et~al.(2014)Sutskever, Vinyals, and
  Le]{sutskever2014sequence}
I.~Sutskever, O.~Vinyals, and Q.~V. Le.
\newblock Sequence to sequence learning with neural networks.
\newblock \emph{Advances in neural information processing systems}, 27, 2014.

\bibitem[Thoppilan et~al.(2022)Thoppilan, De~Freitas, Hall, Shazeer,
  Kulshreshtha, Cheng, Jin, Bos, Baker, Du, et~al.]{thoppilan2022lamda}
R.~Thoppilan, D.~De~Freitas, J.~Hall, N.~Shazeer, A.~Kulshreshtha, H.-T. Cheng,
  A.~Jin, T.~Bos, L.~Baker, Y.~Du, et~al.
\newblock Lamda: Language models for dialog applications.
\newblock \emph{arXiv preprint arXiv:2201.08239}, 2022.

\bibitem[Vaswani et~al.(2017)Vaswani, Shazeer, Parmar, Uszkoreit, Jones, Gomez,
  Kaiser, and Polosukhin]{vaswani2017attention}
A.~Vaswani, N.~Shazeer, N.~Parmar, J.~Uszkoreit, L.~Jones, A.~N. Gomez,
  {\L}.~Kaiser, and I.~Polosukhin.
\newblock Attention is all you need.
\newblock In \emph{Advances in neural information processing systems}, pages
  5998--6008, 2017.

\bibitem[Wang et~al.(2022)Wang, Ruder, and Neubig]{wang2022expanding}
X.~Wang, S.~Ruder, and G.~Neubig.
\newblock Expanding pretrained models to thousands more languages via
  lexicon-based adaptation.
\newblock In \emph{Proceedings of the 60th Annual Meeting of the Association
  for Computational Linguistics (Volume 1: Long Papers)}, pages 863--877, 2022.

\bibitem[Xu et~al.(2018)Xu, Du{\v{s}}ek, Konstas, and Rieser]{xu2018better}
X.~Xu, O.~Du{\v{s}}ek, I.~Konstas, and V.~Rieser.
\newblock Better conversations by modeling, filtering, and optimizing for
  coherence and diversity.
\newblock In \emph{Proceedings of the 2018 Conference on Empirical Methods in
  Natural Language Processing}, pages 3981--3991, 2018.

\bibitem[Zhang et~al.(2020)Zhang, Sun, Galley, Chen, Brockett, Gao, Gao, Liu,
  and Dolan]{zhang2020dialogpt}
Y.~Zhang, S.~Sun, M.~Galley, Y.-C. Chen, C.~Brockett, X.~Gao, J.~Gao, J.~Liu,
  and B.~Dolan.
\newblock Dialogpt: Large-scale generative pre-training for conversational
  response generation.
\newblock In \emph{ACL (demo)}, 2020.

\bibitem[Zhao et~al.(2020)Zhao, Wu, Tao, Xu, Zhao, and Yan]{zhao2020low}
X.~Zhao, W.~Wu, C.~Tao, C.~Xu, D.~Zhao, and R.~Yan.
\newblock Low-resource knowledge-grounded dialogue generation.
\newblock \emph{International Conference on Learning Representations}, 2020.

\end{thebibliography}

\end{document}